\pdfoutput=1
\documentclass[11pt]{article}
\newcommand{\xkg}{\mathrm{\ours}}

\usepackage[]{acl}

\usepackage{times}
\usepackage{latexsym}

\usepackage{booktabs} 
\usepackage{tabularx}   
\usepackage[T1]{fontenc}

\usepackage[utf8]{inputenc}

\usepackage{microtype}

\usepackage{inconsolata}

\usepackage{graphicx}

\newcommand{\ours}{\textsc{xKG}}
\usepackage[skins]{tcolorbox}
\usepackage{xcolor}
\usepackage{amsmath}
\usepackage{multirow}
\usepackage{bbm}
\usepackage{booktabs}
\usepackage{colortbl}
\tcbuselibrary{breakable}    
\usepackage{amsfonts}
\usepackage{fontawesome5}
\usepackage{enumitem}
\usepackage[table]{xcolor} 
\usepackage{makecell}
\definecolor{gaincolor}{HTML}{28a745}
\definecolor{losscolor}{HTML}{dc3545}
\newcommand{\gain}[1]{_{\scriptsize\textcolor{gaincolor}{+#1}}}
\newcommand{\loss}[1]{_{\scriptsize\textcolor{losscolor}{#1}}}

\definecolor{mygray}{gray}{0.9} 
\definecolor{myblue}{HTML}{F0FFFF}
\definecolor{refGreenBG}{HTML}{E6F5E6} 

\newcommand{\textgain}[1]{%
  \setlength{\fboxsep}{1.5pt}
  \colorbox{refGreenBG}{%
    \textbf{\raisebox{1pt}{\( \uparrow \)}}#1
  }%
}

\usepackage{calc}
\newlength\myheight
\newlength\mydepth
\settototalheight\myheight{Xygp}
\title{What Makes AI Research Replicable?\\ Executable Knowledge Graphs as  Scientific Knowledge Representations}

\author{
  \textbf{Yujie Luo}${^{\spadesuit\heartsuit}\footnotemark[1]}$~, 
  \textbf{Zhuoyun Yu}$^{\spadesuit}$\thanks{$\quad$ Equal Contribution.}~,
  \textbf{Xuehai Wang}$^{\spadesuit}$, 
  \textbf{Yuqi Zhu}$^{\spadesuit\heartsuit}$, \\
  \textbf{Ningyu Zhang}$^{\spadesuit\heartsuit}$\thanks{$\quad$ Corresponding Authors.}~, 
  \textbf{Lanning Wei}$^{\clubsuit\heartsuit}$, 
  \textbf{Lun Du}$^{\clubsuit\heartsuit}$, 
  \textbf{Da Zheng}$^{\clubsuit\heartsuit}$\footnotemark[\value{footnote}]~, 
  \textbf{Huajun Chen}$^{\spadesuit\heartsuit}$\footnotemark[\value{footnote}] \\
  $^\spadesuit$Zhejiang University ~$^\clubsuit$Ant Group \\
  $^\heartsuit$Zhejiang University - Ant Group Joint Laboratory of Knowledge Graph 
  \\
  \texttt{\{luo.yj,zhangningyu\}@zju.edu.cn}
}

\begin{document}
\maketitle
\begin{abstract}
Replicating AI research is a crucial yet challenging task for large language model (LLM) agents. Existing approaches often struggle to generate executable code, primarily due to insufficient background knowledge and the limitations of retrieval-augmented generation (RAG) methods, which fail to capture latent technical details hidden in referenced papers. Furthermore, previous approaches tend to overlook valuable implementation-level code signals and lack structured knowledge representations that support multi-granular retrieval and reuse. To overcome these challenges, we propose \textbf{Executable Knowledge Graphs ({\ours})}, a pluggable, paper-centric knowledge base that automatically integrates code snippets and technical insights extracted from scientific literature. When integrated into three agent frameworks with two different LLMs, xKG shows substantial performance gains (10.9\%  with o3-mini) on PaperBench, demonstrating its effectiveness as a general and extensible solution for automated AI research replication\footnote{\url{https://github.com/zjunlp/xKG}.}.
\end{abstract}

\section{Introduction}

The rapid advancement of AI has dramatically accelerated scientific progress, producing thousands of new publications each year \cite{zhao2023survey}. 
However, reproducing these results remains a major bottleneck: many papers omit critical implementation details, code repositories are incomplete or unavailable, and essential background knowledge is scattered across diverse sources \cite{zhao2025autoreproduce,seo2025paper2code,zhou2025reflective,edwards2025rexbench,zhu2025ai,huang2025deep,zhu2025ai,kon2025exp,yan2025lmr}. 
While humans perform the tedious pipeline of reading papers, inspecting code, and collecting background materials to reproduce results, enabling machines to perform the same workflow reliably remains an open challenge \cite{chen2025ai4research}.



\textbf{Why Executable Knowledge Graphs?} 
AI research is hard to replicate and reuse because its knowledge is implicit and fragmented across text, code, and configuration. 
To address this challenge, we propose the \textbf{Executable Knowledge Graph ({\ours})}, a novel, paper-centric knowledge base designed to externalize latent scientific knowledge into a verifiable, executable representation. 

Our {\ours} overcomes the limitations of existing attempts~\cite{DBLP:journals/corr/abs-2505-18705,ou2025automind}, which often stop at coarse-grained knowledge reuse, by modeling scientific literature as hierarchical graphs that ground academic concepts in executable code. 
Unlike conventional KGs, {\ours} captures both conceptual relations and executable components, enabling agents to assemble the precise artifacts needed for faithful reproduction. 
We evaluate {\ours} by integrating it into three distinct agent frameworks: BasicAgent, IterativeAgent, and PaperCoder.
Experiments on PaperBench~\cite{starace2025paperbench} demonstrate consistent and significant performance gains. 
Our {\ours} is built on a fully automated paper-aware pipeline, updatable to stay current with advancing research.

\section{Executable Knowledge Graphs}
\begin{figure*}[t!]
  \centering
  \includegraphics[width=\textwidth]{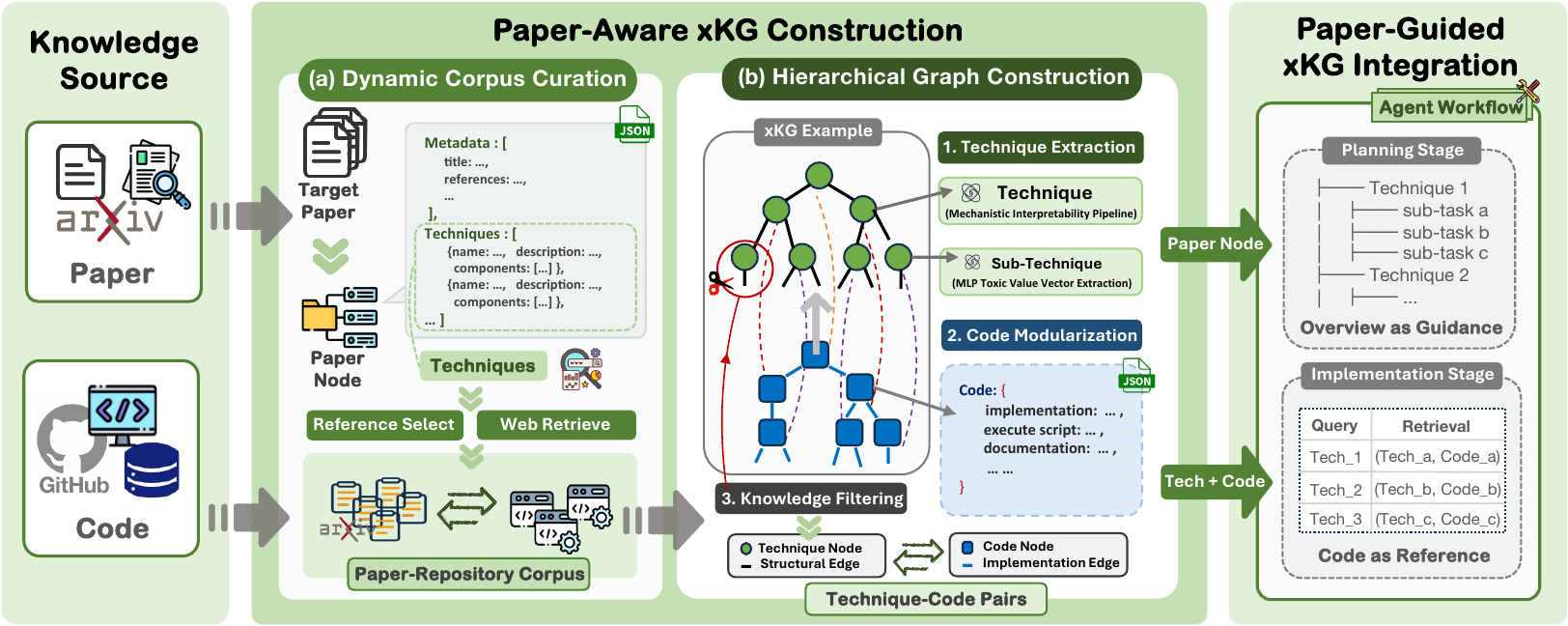}
\caption{\textbf{The Paper-Centric {\ours} Pipeline.} 
\textbf{(1) Construction}: A paper-aware pipeline first curates a corpus of relevant papers and repositories, then extracts, modularizes, and filters it into executable technique-code pairs.
\textbf{(2) Integration}: Our {\ours} guides agents by providing high-level technique overviews and low-level code references.
}
  \label{fig:main}
\end{figure*}
\subsection{Design Formulation}
\label{sec:design}
We model {\ours} as a hierarchical, multi-relational graph $\xkg = (\mathcal{N}, \mathcal{E})$, specifically:
\begin{align}
    \mathcal{N} &= \mathcal{N}_P \cup \mathcal{N}_T \cup \mathcal{N}_C \\
    \mathcal{E} &= \mathcal{E}_{\text{struct}} \cup \mathcal{E}_{\text{impl}} 
\end{align}
We define three types of nodes to capture knowledge at different granularities:
\begin{itemize}[leftmargin=*, noitemsep, topsep=2pt, partopsep=0pt]
    \item \textbf{Paper Node ($n_p$):} Represents a paper as a tuple $n_p = (M_p, \{n_t\}_i, \{n_c\}_j)$, containing metadata $M_p$ (e.g., abstracts, references, etc.), technique nodes $\{n_t\}_i$, and code nodes $\{n_c\}_j$.
    \item \textbf{Technique Node ($n_t$):} A self-contained academic concept $n_t = (D_t, \{n'_{t}\}_k)$ with definition $D_t$ and optional sub-nodes $\{n'_{t}\}_k$, ranging from complete framework to reusable component.
    \item \textbf{Code Node ($n_c$):} An executable unit $n_c = (\sigma, \tau, \delta)$ comprising code implementation $\sigma$, a test script $\tau$, and documentation $\delta$.
\end{itemize}
These nodes are linked by the following edges:
\begin{itemize}[leftmargin=*, noitemsep, topsep=2pt, partopsep=0pt]
    \item \textbf{Structural Edge ($e_{\text{struct}}$):} An edge $(n_{t, i}, n_{t,j})$ indicates an architectural dependency between technique nodes.
    \item \textbf{Implementation Edge ($e_{\text{impl}}$):} An edge $(n_t, n_c)$ linking a technique to its code implementation.
\end{itemize}

Our {\ours} directly links scientific concepts to executable code, yielding more comprehensive knowledge (example in Figure~\ref{fig:main}(b), Appendix \ref{app:xkg-case}).

\subsection{Paper-Aware {\ours} Construction}
 
\subsubsection{Dynamic Corpus Curation}
Our corpus curation is a paper-aware automated pipeline engineered for continuous scalability and cross-domain extensibility, as shown in Figure~\ref{fig:main}(a).
For each paper targeted for reproduction (papers in PaperBench \cite{starace2025paperbench}), {\ours} automates the collection of all prerequisite resources, beginning by employing o4-mini to identify its core \textit{techniques}. 
Centered on these techniques, we first cull its top-ranked references via LLM, and then execute an automated technique-based web retrieval, culminating in a curated set of ten relevant papers for each.
\textbf{Note that we strictly do NOT use any GitHub or third-party reproduction repositories listed in PaperBench’s blacklist to avoid any risk of data leakage}.
All retrieved papers are processed to fetch their \LaTeX\ sources from arXiv and then identify the associated GitHub.
Papers without official repositories are automatically filtered out, resulting in a corpus of \textbf{paper-repository pairs}.
\definecolor{ForestGreen}{rgb}{0.13, 0.55, 0.13}

\newlength{\gaincontentwidth}
\settowidth{\gaincontentwidth}{\small\textbf{$\uparrow$\,10.90}}

\definecolor{refGreenBG}{HTML}{E6F5E6} 
\newcommand{\hlgainsmall}[1]{{%
  \setlength{\fboxsep}{1pt}%
  \colorbox{refGreenBG}{%
    \makebox[\gaincontentwidth][l]{\strut\small\textbf{$\uparrow$\,#1}}%
  }%
}}
\newcommand{\hllosssmall}[1]{{%
  \setlength{\fboxsep}{1pt}%
  \colorbox{red!20}{%
    \makebox[\gaincontentwidth][l]{\strut\small\textbf{$\downarrow$\,#1}}%
  }%
}}


\definecolor{refBlueBG}{HTML}{E5F0FF} 
\newcommand{\headerours}{{%
  \setlength{\fboxsep}{5pt}%
  \colorbox{refBlueBG}{+\ours}%
}}

\begin{table*}[htp]
    \centering
    \setlength{\tabcolsep}{0.7mm}
    \resizebox{\textwidth}{!}{
    \renewcommand{\arraystretch}{1.3}
    \begin{tabular}{c c cc cc cc cc cc cc}
    \toprule[1.2pt]
    \multirow{3}{*}[1ex]{\textbf{Method}} & \multirow{3}{*}[1ex]{\textbf{Model}} & \multicolumn{2}{c}{\textbf{MU-DPO}} & \multicolumn{2}{c}{\textbf{TTA-FP}}  & \multicolumn{2}{c}{\textbf{One-SBI}}  & \multicolumn{2}{c}{\textbf{CFG}}  & \multicolumn{2}{c}{\textbf{FRE}} & \multicolumn{2}{c}{\textbf{Average}}\\[0.2ex]

    &  & \hspace*{0.2em}vanilla\hspace*{0.2em} & \hspace*{0.2em}\headerours\hspace*{0.2em} & \hspace*{0.2em}vanilla\hspace*{0.2em} & \hspace*{0.2em}\headerours\hspace*{0.2em} & \hspace*{0.2em}vanilla\hspace*{0.2em} & \hspace*{0.2em}\headerours\hspace*{0.2em}  & \hspace*{0.2em}vanilla\hspace*{0.2em} & \hspace*{0.2em}\headerours\hspace*{0.2em} & \hspace*{0.2em}vanilla\hspace*{0.2em} & \hspace*{0.2em}\headerours\hspace*{0.2em} & \hspace*{0.2em}vanilla\hspace*{0.2em} & \hspace*{0.2em}\headerours\hspace*{0.2em}\\

    \midrule[1pt]
    \multirow{2}{*}{BasicAgent} 
    & \texttt{o3-mini} & 12.96 & $37.22\gain{24.26}$  & 22.63 & $27.26\gain{4.63}$  & 18.24 & $20.82\gain{2.58}$  & 20.82 & $22.86\gain{2.04}$  & 14.82 & $14.67\loss{-0.15}$   
    & 17.89 & 24.57\,\hlgainsmall{6.68} \\
    
    & \texttt{DS-R1}   & 33.05 & $39.14\gain{6.09}$   & 40.55 & $39.14\loss{-1.41}$ & 17.22 & $24.49\gain{7.27}$  & 31.56 & $33.97\gain{2.41}$  & 17.08 & $21.38\gain{4.30}$  
    & 27.89 & 31.62\,\hlgainsmall{3.73} \\
    \addlinespace
    
    \multirow{2}{*}{IterativeAgent} 
    & \texttt{o3-mini} & 22.22 & $43.70\gain{21.48}$  & 21.38 & $36.28\gain{14.90}$ & 28.77 & $23.91\loss{-4.86}$ & 31.28 & $29.15\loss{-2.13}$ & 19.35 & $26.50\gain{7.15}$  
    & 24.60 & 31.91\,\hlgainsmall{7.31} \\
    
    & \texttt{DS-R1}   & 16.20 & $47.40\gain{31.20}$  & 31.19 & $31.78\gain{0.59}$  & 31.09 & $26.57\loss{-4.52}$ & 35.30 & $38.44\gain{3.14}$  & 21.32 & $31.89\gain{10.57}$ 
    & 27.02 & 35.22\,\hlgainsmall{8.20} \\
    \addlinespace
    
    \multirow{2}{*}{PaperCoder} 
    & \texttt{o3-mini} & 23.15 & $46.48\gain{23.33}$  & 45.70 & $53.99\gain{8.29}$  & 52.48 & $52.08\loss{-0.40}$ & 50.37 & $63.13\gain{12.76}$ & 39.84 & $50.36\gain{10.52}$ 
    & 42.31 & 53.21\,\hlgainsmall{10.90} \\
    
    & \texttt{DS-R1}   & 43.24 & $49.26\gain{6.02}$   & 43.26 & $59.19\gain{15.93}$ & 51.18 & $73.03\gain{21.85}$ & 61.12 & $60.68\loss{-0.44}$ & 62.37 & $59.53\loss{-2.84}$ 
    & 52.23 & 60.34\,\hlgainsmall{8.11} \\
    
    \bottomrule[1.2pt]
    \end{tabular}
    }
    \caption{\textbf{Main results on PaperBench Code-Dev.} We evaluate on the official \texttt{lite} subset of PaperBench (details in Table \ref{tab:paperbench-lite}). All results are the \texttt{best@3} \textit{Replication Score} (\%) to mitigate task stochasticity and potential tool failures.}
    \label{tab:main}
\end{table*}

\subsubsection{Hierarchical KG Construction}
Based on the corpus, we then construct the {\ours} with the automated steps shown in Figure~\ref{fig:main}(b):
\begin{itemize}[leftmargin=*]
    \item \textbf{Step 1: Technique Extraction.} We first use o4-mini to deconstruct the paper's decomposable methodology into a preliminary hierarchical tree of Technique Nodes $\mathcal{N}_T$ linked by Structural Edges $e_{\text{struct}}$(details in Appendix~\ref{app:granularity}). 
    Subsequently, we utilize RAG\footnote{We employ text-embedding-3-small for embedding similarity throughout all stages of xKG construction.} to enrich each node by retrieving relevant text from the paper to form a comprehensive definition $D_t$. 
    This step yields a set of detailed techniques yet may contain noise.

    \item \textbf{Step 2: Code Modularization.}  
    For each technique $n_t$, its definition is used as a query to retrieve code snippets from the repository via embedding similarity. We then employ o4-mini to synthesize these snippets into candidate Code Nodes $n_c$, each includes the implementation $\sigma$, test script $\tau$, and documentation $\delta$. These nodes then undergo an iterative \textit{self-debugging} loop to ensure executability, ultimately producing a set of \textbf{executable Code Nodes} $\mathcal{N}_c$ with associated technique-linking Implementation Edges $e_{\text{impl}}$.

    \item \textbf{Step 3: Knowledge Filtering.}
    We formalize a simple yet powerful verification principle: a technique $n_t$  is considered valuable only if it can be grounded in executable code.
    Therefore, any technique for which \textbf{Step 2} failed to retrieve relevant code snippets is pruned from the {\ours}.
    This filtering process ensures that only techniques with proven, practical value populate the final {\ours}, eliminating the noise and overly granular nodes introduced in \textbf{Step 1}.
\end{itemize}

Finally, we construct {\ours} from 42 curated papers, totaling 591,145 tokens. \textbf{Our {\ours} construction is built upon a paper-aware pipeline designed for continuous evolution as new literature emerges}, as detailed in Appendix \ref{app:xkg_evolution}.

\subsection{Paper-Guided {\ours} Integration}


{\ours} can be seamlessly integrated into a practical reproduction workflow, where LLM agents utilize it at two stages (Figure~\ref{fig:main}, right).
For \textbf{high-level planning}, the agent fetches the target paper's Paper Node (without Code Nodes) to grasp its core techniques and overall structure. 
During \textbf{low-level implementation}, the agent queries {\ours} for (Technique, Code) pairs directly relevant to the target paper.
These two steps can be supplied either as callable tools for ReAct agents or as pluggable components of fixed-workflow agents.
Crucially, all retrieved candidates are processed by a final \textbf{LLM-based Verifier} (o4-mini), acting as a final quality gate to ensure the retrieved knowledge is both technically relevant and implementable.

\section{Experiments}
\subsection{Settings}

We evaluate {\ours} on the lite collection of PaperBench Code-Dev \cite{starace2025paperbench}, a benchmark for repository-level paper reproduction from scratch, scored by an o3-mini-based evaluator using a weighted, tree-structured rubric. 
We integrate {\ours} into BasicAgent (a ReAct-style agent), IterativeAgent (adds a self-improvement loop), both with a one-hour runtime limit, and PaperCoder(a powerful agent tailored for repository-level paper reproduction.). 
See Appendix \ref{app:setup} for more details.

\subsection{Main Results} 

As shown in Table~\ref{tab:main}, {\ours} achieves substantial performance gains across diverse agent frameworks and LLM backbones. 
On the general ReAct-style IterativeAgent with DeepSeek-R1, {\ours} delivers a performance improvement of \textgain{8.20\%}. 
The effectiveness of {\ours} is further highlighted by the \textgain{10.90\%} improvement achieved with PaperCoder powered with o3-mini, underscoring its broad applicability from simpler agents to more advanced ones.
Notably, the impact of {\ours} is also highly paper-dependent.
While BasicAgent with o3-mini achieves a remarkable 24.26\% performance gain on MU-DPO, the same configuration yields only a 2.58\% improvement on One-SBI and even a 0.15\% drop on the FRE task, revealing a critical dependency on the target paper (details in Appendix \ref{app:app_experiments}).

\subsection{Further Analysis} 
\begin{table}[h]
\centering
\small 
\renewcommand{\arraystretch}{1.2}
\begin{tabular*}{\columnwidth}{l @{\extracolsep{\fill}} cc} 
\toprule[1.1pt]
\textbf{Method} & \textbf{Score (\%)} & \textbf{Drop ($\nabla$)} \\
\midrule[1pt]
\ours (Full)        & 53.21 & -    \\
\midrule
w/o Paper Node      & 51.08 & 2.13 \\
w/o Code Node       & 48.65 & 4.56 \\
w/o Technique Node  & 52.16 & 1.05 \\
\bottomrule[1.1pt]
\end{tabular*}
\caption{Ablation study on {\ours} node types.}
\label{tab:ablation}
\end{table}

\paragraph{Code-based structured knowledge aids AI research replication.}
As shown in Table~\ref{tab:ablation}, our ablation study conducted on PaperCoder with o3-mini reveals that removing any component degrades performance. 
The most significant drop occurs when removing \textbf{Code Nodes}, decreasing the score by 4.56\% (53.21\% $\rightarrow$ 48.65\%), suggesting that LLM agents benefit immensely from fine-grained knowledge, with executable code being the most critical component.
Ablating \textbf{Paper Nodes} yields a substantial degradation of 2.13\%, highlighting the value of a structural overview of the target paper.
In contrast, omitting \textbf{Technique Nodes} results in a modest 1.05\% drop, since the function of each technique is already captured by the Code Nodes, rendering the explicit description redundant.

\paragraph{Successful reproduction hinges on retrieved code quality.}
Building on the above, we conduct a further analysis of how Code Nodes influence performance. 
We analyze the impact of code quality on PaperCoder (o3-mini), comparing four settings with 3 runs each: w/o Code (no code), + Raw Code (raw snippets), + Rewrite (LLM-rewritten but unverified), and + Verify (LLM-rewritten \& verified).

\begin{figure}[t!]
  \centering
  \includegraphics[width=0.9\linewidth]{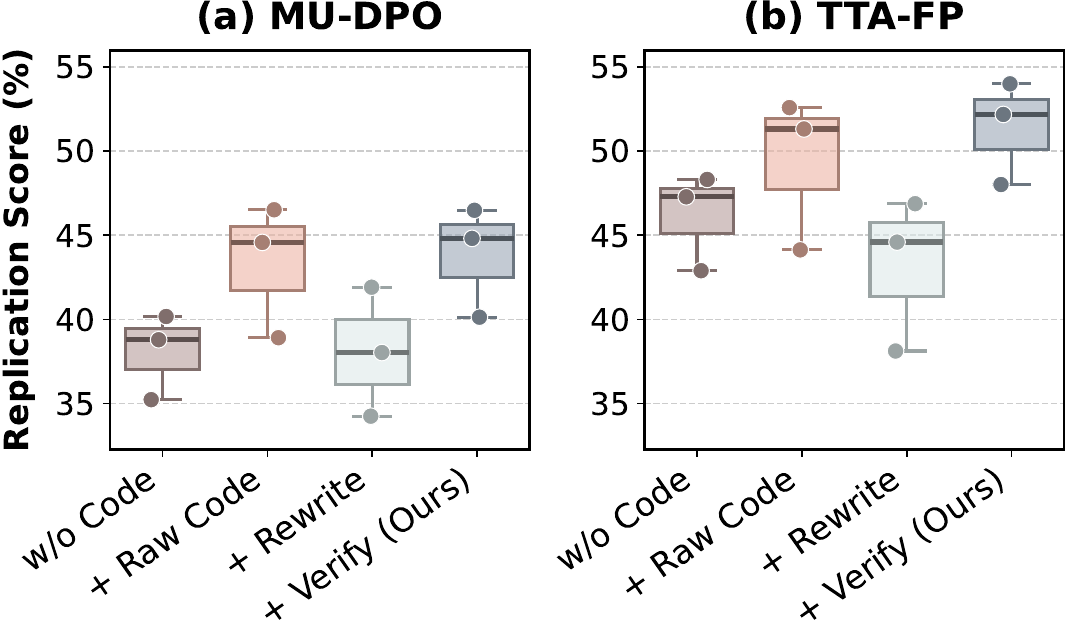}
  \caption{Further study on Code Node quality.}
  \label{fig:ablation}
\end{figure}

\begin{table}[t]
\centering
\small 
\renewcommand{\arraystretch}{1.2}
\begin{tabular}{lccc} 
\toprule[1.1pt]
& \textbf{Techs.} & \textbf{Codes} & \textbf{Tech-Code Pairs} \\
\midrule 
\textbf{Valid Rate (\%)} & 89.44 & 100.00 & 74.51 \\
\bottomrule[1.1pt]
\end{tabular}
\caption{Human evaluation of xKG quality.}
\label{tab:quality_eval_compact}
\end{table}

\begin{figure}[t!]
  \centering
  \includegraphics[width=0.95\linewidth]{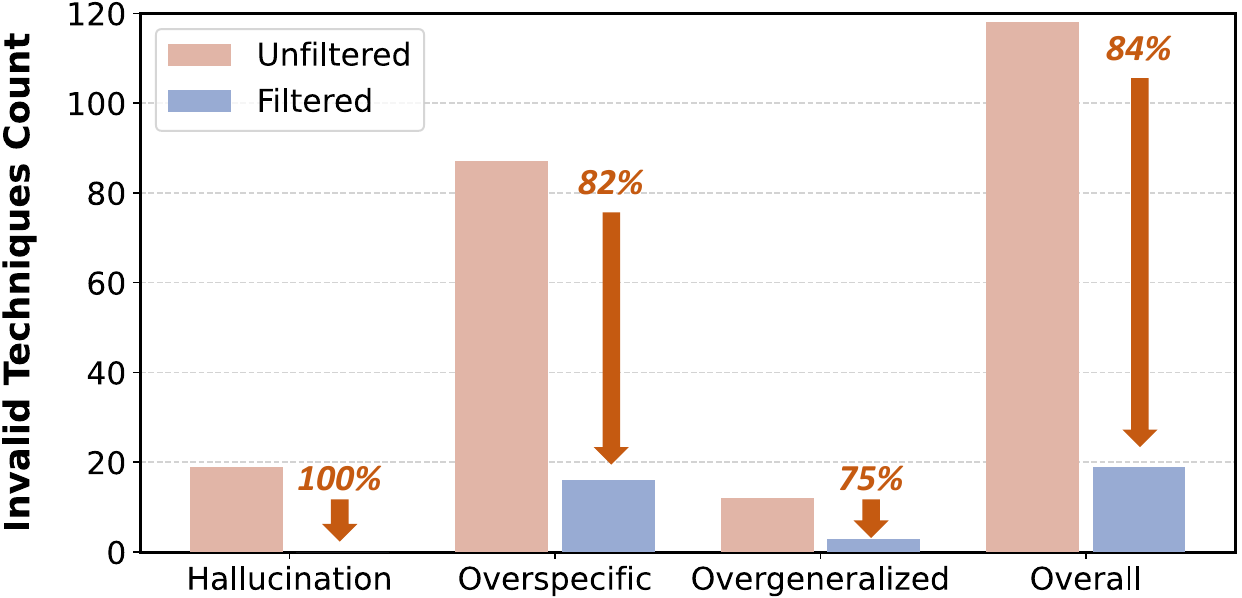}
  \caption{Comparison of invalid Technique Nodes in {\ours} before and after knowledge filtering.}
  \label{fig:compare}
\end{figure}

As illustrated in Figure~\ref{fig:ablation}, our full method excels in score and stability. Notably, even incorporating raw code snippets (\textit{+ Raw Code}) significantly improves performance, validating that our method effectively localizes necessary code. The failure of \textit{+ Rewrite} ablation reveals a key insight: \textbf{agents are misled by well-formatted, semantically similar yet technically irrelevant knowledge}, a phenomenon consistent with previous findings \cite{DBLP:journals/corr/abs-2404-03302}. Our LLM Verifier mitigates this by prioritizing technical relevance over semantic similarity, filtering out distracting irrelevant information, thus boosting the score improvement varying from 5.75 to 8.20 percentage points.


\paragraph{Automated Construction architects robust {\ours}.}

Human evaluation (Table~\ref{tab:quality_eval_compact}) confirms the high quality of {\ours} with 89.44\% of Techniques representing self-contained academic concepts, all Code Nodes executable, and 74.51\% of Tech-Code pairs exactly matched. 
During {\ours} construction, the Knowledge Filtering step mitigates LLM errors (e.g., overspecificity, hallucinations) by pruning techniques with no code retrieved (Figure~\ref{fig:compare}). 
Concurrently, a self-debugging loop achieves 100\% code executability, up from an initial 52.38\%. 
Tech-Code mismatches primarily stem from broader code snippets with little impact on application. 
See Appendix~\ref{app:app_xkg} for more details of constructed {\ours}.


\paragraph{xKG Transforms Agents from Scaffolding to Implementation.}

\begin{figure}[t!]
  \centering
  \includegraphics[width=1\linewidth]{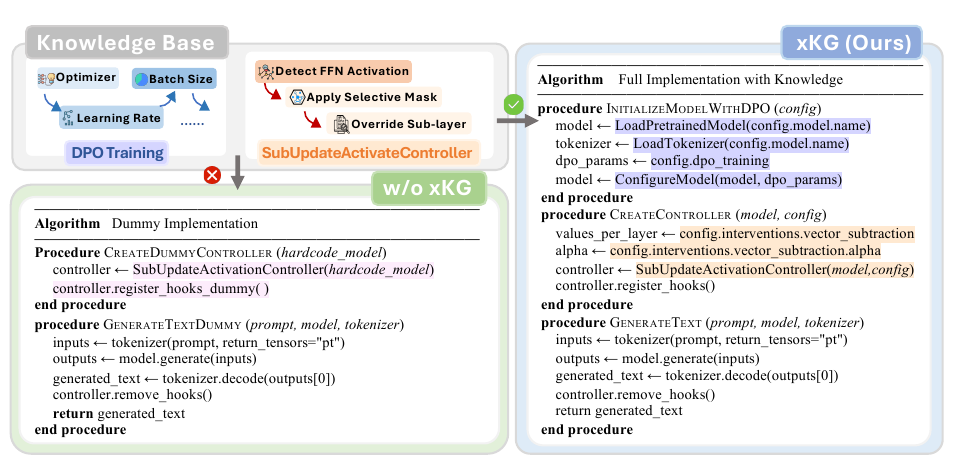}
  \caption{Case Study on MU-DPO. Highlights illustrating agent implementation contrast with/without {\ours}.}

  \label{fig:case}
\end{figure}

To understand the mechanism behind the performance gains, we conduct a case study on the MU-DPO paper (Figure ~\ref{fig:case}). 
We notice that {\ours} enriches information granularity, allowing agents to generate critical details accurately, and improves modular implementation capability, enabling agents to reuse code for functionally correct implementations, as highlighted in Figure \ref{fig:case}.
This case reveals that {\ours} transforms agents from dummy scaffolders to substantive implementers, arming them with both the precise method details for accurate planning and the verified, modular reference code for robust implementation.


\section{Conclusion} 
We introduce Executable Knowledge Graphs ({\ours}) to make implicit research knowledge modular and executable, boosting agent replication performance. 
Looking forward, we envision {\ours} serving not only as a dynamic knowledge base, but also as a flexible instrument to accelerate AI Research with improved efficiency and verifiability.


\section*{Limitations} 

This work has several limitations. 
First, the PaperBench task exhibits high variance and is costly to evaluate. 
Although we report results across multiple papers and conduct experiments, due to funding constraints, we only perform experiments on the \texttt{lite} collection of PaperBench Code-Dev. 
Second, for emerging domains, there may be no available reference papers at all, which limits the applicability of our approach to scenarios where some baseline references exist.
Finally, while the code-based knowledge organization we propose may have the potential to transfer to similar tasks, exploring this remains future work \cite{nathani2025mlgym,chan2024mle,toledo2025ai,jia2025towards,miao2025paper2agent}.

During our work, we found another project with a similar name, ExeKG \cite{zheng2022exekg,zheng2022executable,zhou2022towards}.
However, our approach differs fundamentally in the organization of the knowledge base — we adopt a much simpler structure of nodes and edges.
Moreover, the problems addressed are entirely distinct: our focus is on paper replication tasks.
We hold deep respect for the pioneering efforts of the ExeKG authors.


\section*{Acknowledgement}
We would like to express sincere gratitude to the  reviewers for their thoughtful and constructive feedback. This work was supported by the National Natural Science Foundation of China (No. 62576307, No. NSFCU23B2055, No. NSFCU19B2027), the Fundamental Research Funds for the Central Universities (226-2023-00138), Yongjiang Talent Introduction Programme (2021A-156-G), and Information Technology Center and State Key Lab of CAD\&CG, Zhejiang University. This work was supported by Ant Group and Zhejiang University - Ant Group Joint Laboratory of Knowledge Graph.

\bibliography{custom}

\begin{thebibliography}{27}
\providecommand{\natexlab}[1]{#1}

\bibitem[{Chan et~al.(2024)Chan, Chowdhury, Jaffe, Aung, Sherburn, Mays, Starace, Liu, Maksin, Patwardhan et~al.}]{chan2024mle}
Jun~Shern Chan, Neil Chowdhury, Oliver Jaffe, James Aung, Dane Sherburn, Evan Mays, Giulio Starace, Kevin Liu, Leon Maksin, Tejal Patwardhan, et~al. 2024.
\newblock Mle-bench: Evaluating machine learning agents on machine learning engineering.
\newblock \emph{arXiv preprint arXiv:2410.07095}.

\bibitem[{Chen et~al.(2025)Chen, Yang, Qin, Liu, Yan, Guan, Peng, Ji, Li, Hu et~al.}]{chen2025ai4research}
Qiguang Chen, Mingda Yang, Libo Qin, Jinhao Liu, Zheng Yan, Jiannan Guan, Dengyun Peng, Yiyan Ji, Hanjing Li, Mengkang Hu, et~al. 2025.
\newblock Ai4research: A survey of artificial intelligence for scientific research.
\newblock \emph{arXiv preprint arXiv:2507.01903}.

\bibitem[{Edwards et~al.(2025)Edwards, Lee, Mao, Qin, Schuster, and Kim}]{edwards2025rexbench}
Nicholas Edwards, Yukyung Lee, Yujun~Audrey Mao, Yulu Qin, Sebastian Schuster, and Najoung Kim. 2025.
\newblock Rexbench: Can coding agents autonomously implement ai research extensions?
\newblock \emph{arXiv preprint arXiv:2506.22598}.

\bibitem[{Frans et~al.(2024)Frans, Park, Abbeel, and Levine}]{DBLP:conf/icml/FransPAL24}
Kevin Frans, Seohong Park, Pieter Abbeel, and Sergey Levine. 2024.
\newblock \href {https://openreview.net/forum?id=a6wCNfIj8E} {Unsupervised zero-shot reinforcement learning via functional reward encodings}.
\newblock In \emph{Forty-first International Conference on Machine Learning, {ICML} 2024, Vienna, Austria, July 21-27, 2024}. OpenReview.net.

\bibitem[{Gl{\"{o}}ckler et~al.(2024)Gl{\"{o}}ckler, Deistler, Weilbach, Wood, and Macke}]{DBLP:conf/icml/GlocklerDWWM24}
Manuel Gl{\"{o}}ckler, Michael Deistler, Christian~Dietrich Weilbach, Frank Wood, and Jakob~H. Macke. 2024.
\newblock \href {https://openreview.net/forum?id=DL79HYCFFq} {All-in-one simulation-based inference}.
\newblock In \emph{Forty-first International Conference on Machine Learning, {ICML} 2024, Vienna, Austria, July 21-27, 2024}. OpenReview.net.

\bibitem[{Huang et~al.(2025)Huang, Chen, Zhang, Li, Zhou, Fang, Yang, Li, Shang, Xu et~al.}]{huang2025deep}
Yuxuan Huang, Yihang Chen, Haozheng Zhang, Kang Li, Huichi Zhou, Meng Fang, Linyi Yang, Xiaoguang Li, Lifeng Shang, Songcen Xu, et~al. 2025.
\newblock Deep research agents: A systematic examination and roadmap.
\newblock \emph{arXiv preprint arXiv:2506.18096}.

\bibitem[{Jia et~al.(2025)Jia, Qian, Tong, Wu, Chen, and Wei}]{jia2025towards}
Hangyi Jia, Yuxi Qian, Hanwen Tong, Xinhui Wu, Lin Chen, and Feng Wei. 2025.
\newblock Towards adaptive ml benchmarks: Web-agent-driven construction, domain expansion, and metric optimization.
\newblock \emph{arXiv preprint arXiv:2509.09321}.

\bibitem[{Kon et~al.(2025)Kon, Liu, Zhu, Ding, Peng, Xing, Huang, Qiu, Srinivasa, Lee et~al.}]{kon2025exp}
Patrick Tser~Jern Kon, Jiachen Liu, Xinyi Zhu, Qiuyi Ding, Jingjia Peng, Jiarong Xing, Yibo Huang, Yiming Qiu, Jayanth Srinivasa, Myungjin Lee, et~al. 2025.
\newblock Exp-bench: Can ai conduct ai research experiments?
\newblock \emph{arXiv preprint arXiv:2505.24785}.

\bibitem[{Lee et~al.(2024)Lee, Bai, Pres, Wattenberg, Kummerfeld, and Mihalcea}]{DBLP:conf/icml/LeeBPWKM24}
Andrew Lee, Xiaoyan Bai, Itamar Pres, Martin Wattenberg, Jonathan~K. Kummerfeld, and Rada Mihalcea. 2024.
\newblock \href {https://openreview.net/forum?id=dBqHGZPGZI} {A mechanistic understanding of alignment algorithms: {A} case study on {DPO} and toxicity}.
\newblock In \emph{Forty-first International Conference on Machine Learning, {ICML} 2024, Vienna, Austria, July 21-27, 2024}. OpenReview.net.

\bibitem[{Miao et~al.(2025)Miao, Davis, Pritchard, and Zou}]{miao2025paper2agent}
Jiacheng Miao, Joe~R Davis, Jonathan~K Pritchard, and James Zou. 2025.
\newblock Paper2agent: Reimagining research papers as interactive and reliable ai agents.
\newblock \emph{arXiv preprint arXiv:2509.06917}.

\bibitem[{Nathani et~al.(2025)Nathani, Madaan, Roberts, Bashlykov, Menon, Moens, Budhiraja, Magka, Vorotilov, Chaurasia et~al.}]{nathani2025mlgym}
Deepak Nathani, Lovish Madaan, Nicholas Roberts, Nikolay Bashlykov, Ajay Menon, Vincent Moens, Amar Budhiraja, Despoina Magka, Vladislav Vorotilov, Gaurav Chaurasia, et~al. 2025.
\newblock Mlgym: A new framework and benchmark for advancing ai research agents.
\newblock \emph{arXiv preprint arXiv:2502.14499}.

\bibitem[{Niu et~al.(2024)Niu, Miao, Chen, Wu, and Zhao}]{DBLP:conf/icml/NiuMCWZ24}
Shuaicheng Niu, Chunyan Miao, Guohao Chen, Pengcheng Wu, and Peilin Zhao. 2024.
\newblock \href {https://openreview.net/forum?id=qz1Vx1v9iK} {Test-time model adaptation with only forward passes}.
\newblock In \emph{Forty-first International Conference on Machine Learning, {ICML} 2024, Vienna, Austria, July 21-27, 2024}. OpenReview.net.

\bibitem[{Ou et~al.(2025)Ou, Luo, Zheng, Wei, Qiao, Zhang, Zheng, Chen, and Zhang}]{ou2025automind}
Yixin Ou, Yujie Luo, Jingsheng Zheng, Lanning Wei, Shuofei Qiao, Jintian Zhang, Da~Zheng, Huajun Chen, and Ningyu Zhang. 2025.
\newblock Automind: Adaptive knowledgeable agent for automated data science.
\newblock \emph{arXiv preprint arXiv:2506.10974}.

\bibitem[{Sanchez et~al.(2024)Sanchez, Spangher, Fan, Levi, and Biderman}]{DBLP:conf/icml/SanchezSFLB24}
Guillaume Sanchez, Alexander Spangher, Honglu Fan, Elad Levi, and Stella Biderman. 2024.
\newblock \href {https://openreview.net/forum?id=RiM3cl9MdK} {Stay on topic with classifier-free guidance}.
\newblock In \emph{Forty-first International Conference on Machine Learning, {ICML} 2024, Vienna, Austria, July 21-27, 2024}. OpenReview.net.

\bibitem[{Seo et~al.(2025)Seo, Baek, Lee, and Hwang}]{seo2025paper2code}
Minju Seo, Jinheon Baek, Seongyun Lee, and Sung~Ju Hwang. 2025.
\newblock Paper2code: Automating code generation from scientific papers in machine learning.
\newblock \emph{arXiv preprint arXiv:2504.17192}.

\bibitem[{Starace et~al.(2025)Starace, Jaffe, Sherburn, Aung, Chan, Maksin, Dias, Mays, Kinsella, Thompson et~al.}]{starace2025paperbench}
Giulio Starace, Oliver Jaffe, Dane Sherburn, James Aung, Jun~Shern Chan, Leon Maksin, Rachel Dias, Evan Mays, Benjamin Kinsella, Wyatt Thompson, et~al. 2025.
\newblock Paperbench: Evaluating ai's ability to replicate ai research.
\newblock \emph{arXiv preprint arXiv:2504.01848}.

\bibitem[{Tang et~al.(2025)Tang, Xia, Li, and Huang}]{DBLP:journals/corr/abs-2505-18705}
Jiabin Tang, Lianghao Xia, Zhonghang Li, and Chao Huang. 2025.
\newblock \href {https://doi.org/10.48550/ARXIV.2505.18705} {Ai-researcher: Autonomous scientific innovation}.
\newblock \emph{CoRR}, abs/2505.18705.

\bibitem[{Toledo et~al.(2025)Toledo, Hambardzumyan, Josifoski, Hazra, Baldwin, Audran-Reiss, Kuchnik, Magka, Jiang, Lupidi et~al.}]{toledo2025ai}
Edan Toledo, Karen Hambardzumyan, Martin Josifoski, Rishi Hazra, Nicolas Baldwin, Alexis Audran-Reiss, Michael Kuchnik, Despoina Magka, Minqi Jiang, Alisia~Maria Lupidi, et~al. 2025.
\newblock Ai research agents for machine learning: Search, exploration, and generalization in mle-bench.
\newblock \emph{arXiv preprint arXiv:2507.02554}.

\bibitem[{Wu et~al.(2024)Wu, Xie, Chen, Zhu, Zhang, and Xiao}]{DBLP:journals/corr/abs-2404-03302}
Siye Wu, Jian Xie, Jiangjie Chen, Tinghui Zhu, Kai Zhang, and Yanghua Xiao. 2024.
\newblock \href {https://doi.org/10.48550/ARXIV.2404.03302} {How easily do irrelevant inputs skew the responses of large language models?}
\newblock \emph{CoRR}, abs/2404.03302.

\bibitem[{Yan et~al.(2025)Yan, Li, Luo, Wang, Li, Jing, He, Wu, Michalopoulos, Zhang et~al.}]{yan2025lmr}
Shuo Yan, Ruochen Li, Ziming Luo, Zimu Wang, Daoyang Li, Liqiang Jing, Kaiyu He, Peilin Wu, George Michalopoulos, Yue Zhang, et~al. 2025.
\newblock Lmr-bench: Evaluating llm agent's ability on reproducing language modeling research.
\newblock \emph{arXiv preprint arXiv:2506.17335}.

\bibitem[{Zhao et~al.(2023)Zhao, Zhou, Li, Tang, Wang, Hou, Min, Zhang, Zhang, Dong et~al.}]{zhao2023survey}
Wayne~Xin Zhao, Kun Zhou, Junyi Li, Tianyi Tang, Xiaolei Wang, Yupeng Hou, Yingqian Min, Beichen Zhang, Junjie Zhang, Zican Dong, et~al. 2023.
\newblock A survey of large language models.
\newblock \emph{arXiv preprint arXiv:2303.18223}, 1(2).

\bibitem[{Zhao et~al.(2025)Zhao, Sang, Li, Shi, Zhao, Wang, Zhang, Han, Liu, and Sun}]{zhao2025autoreproduce}
Xuanle Zhao, Zilin Sang, Yuxuan Li, Qi~Shi, Weilun Zhao, Shuo Wang, Duzhen Zhang, Xu~Han, Zhiyuan Liu, and Maosong Sun. 2025.
\newblock Autoreproduce: Automatic ai experiment reproduction with paper lineage.
\newblock \emph{arXiv preprint arXiv:2505.20662}.

\bibitem[{Zheng et~al.(2022{\natexlab{a}})Zheng, Zhou, Zhou, Soylu, and Kharlamov}]{zheng2022executable}
Zhuoxun Zheng, Baifan Zhou, Dongzhuoran Zhou, Ahmet Soylu, and Evgeny Kharlamov. 2022{\natexlab{a}}.
\newblock Executable knowledge graph for transparent machine learning in welding monitoring at bosch.
\newblock In \emph{Proceedings of the 31st ACM International Conference on Information \& Knowledge Management}, pages 5102--5103.

\bibitem[{Zheng et~al.(2022{\natexlab{b}})Zheng, Zhou, Zhou, Soylu, and Kharlamov}]{zheng2022exekg}
Zhuoxun Zheng, Baifan Zhou, Dongzhuoran Zhou, Ahmet Soylu, and Evgeny Kharlamov. 2022{\natexlab{b}}.
\newblock Exekg: Executable knowledge graph system for user-friendly data analytics.
\newblock In \emph{Proceedings of the 31st ACM international conference on information \& knowledge management}, pages 5064--5068.

\bibitem[{Zhou et~al.(2022)Zhou, Zhou, Zheng, Tan, Kostylev, and Kharlamov}]{zhou2022towards}
Dongzhuoran Zhou, Baifan Zhou, Zhuoxun Zheng, Zhipeng Tan, Egor~V Kostylev, and Evgeny Kharlamov. 2022.
\newblock Towards executable knowledge graph translation.
\newblock In \emph{ISWC (Posters/Demos/Industry)}.

\bibitem[{Zhou et~al.(2025)Zhou, Yao, Du, Wei, and Zheng}]{zhou2025reflective}
Mingyang Zhou, Quanming Yao, Lun Du, Lanning Wei, and Da~Zheng. 2025.
\newblock Reflective paper-to-code reproduction enabled by fine-grained verification.
\newblock \emph{arXiv preprint arXiv:2508.16671}.

\bibitem[{Zhu et~al.(2025)Zhu, Xie, Weng, Wu, Lin, Yang, and Zhang}]{zhu2025ai}
Minjun Zhu, Qiujie Xie, Yixuan Weng, Jian Wu, Zhen Lin, Linyi Yang, and Yue Zhang. 2025.
\newblock Ai scientists fail without strong implementation capability.
\newblock \emph{arXiv preprint arXiv:2506.01372}.

\end{thebibliography}




\clearpage
\appendix

\section{Experimental Details}
\label{app:setup}
\subsection{Benchmarks}
\label{app:benchmarks}

The original PaperBench benchmark~\cite{starace2025paperbench}, featuring publicly available tasks and evaluation rubrics, is designed to evaluate the ability of AI agents to reproduce AI research from scratch.

As full-scale evaluation is both computationally expensive and time-consuming, the authors introduced a lightweight variant, PaperBench Code-Dev, which focuses solely on code development—assessing implementation correctness without requiring code execution or result verification.

In our study, we adopt the pre-defined lite subset of PaperBench Code-Dev provided in the official repository, spanning diverse AI Research domains such as machine learning, reinforcement learning, and natural language processing. We further analyze and categorize the research domains and related techniques involved in the target papers, as detailed in Table \ref{tab:paperbench-domain}.

Evaluation follows a structured hierarchical rubric co-developed with the original authors, and an LLM-based evaluator (\texttt{o3-mini}) aggregates the final scores using a weighted binary criteria tree. Specific details about the papers and their evaluation nodes are listed in Table \ref{tab:paperbench-lite}.

Furthermore, since PaperBench shows that \textsc{BasicAgent} and \textsc{IterativeAgent} achieve little performance improvement beyond one hour, we cap their execution time at one hour for efficiency and cost control.

\subsection{Configuration}
\label{app:hyperparameters_researchkg}
\label{app:prompts}
The configuration of our \ours{} framework comprises both hyperparameters and prompts. The hyperparameters are managed via a central \texttt{config.yaml} file, which is organized into modules for Code-RAG, Paper-RAG, and Knowledge Graph Retrieval. 
We summarize the key parameters for each module in Tables~\ref{tab:hyperparams_coderag}-\ref{tab:hyperparams_global}. In addition, the specific prompts designed in our system are detailed in Appendix~\ref{app:prompts_details}.

\subsection{Cost Evaluation}
Our xKG construction pipeline leverages OpenAI's o4-mini via third-party API, chosen for its strong cost-effectiveness and robust text/code capabilities. We quantify the cost of this process in Table \ref{tab:construction-cost}.

As shown, for an average cost of about \$0.73 per paper, our pipeline transforms a scientific publication into an executable knowledge graph. The most significant cost driver is the iterative \textit{self-debugging} loop of the Code Modularization stage, simultaneously serving as the critical quality-assurance mechanism for our code nodes. The key advantage of xKG lies in this one-time investment: \textbf{each processed paper is converted into a durable and reusable knowledge resource}.

\section{Analysis of constructed {\ours}}

\label{app:app_xkg}

\subsection{Self-Evolution of {\ours}}
\label{app:xkg_evolution}

 
{\ours} is not a static knowledge base but a flexible framework that automates the collection, extraction, and validation of knowledge in a paper-aware way. To illustrate this, we select two additional target papers from PaperBench, \textit{bridging-data-gaps} and \textit{sample-specific-masks}, to demonstrate the dynamic evolution of {\ours}.

The expansion from a 42-paper corpus (Table~\ref{tab:xkg_domain}) to a 56-paper version (Table~\ref{tab:xkg_domain_updated}), curated from publicly available research, highlights the autonomous evolution of {\ours}. From the specific replication scores on these tasks (Table~\ref{tab:xKG-results}), we can further observe that {\ours} maintains sustained performance gains on newly introduced paper replication tasks.

When new target papers are introduced, our Corpus Curation module automatically gathers relevant literature and completes the full construction and verification cycle. This adaptability substantiates our claim that {\ours} is an adaptive system, effectively handling scenarios like fetching updated research or leveraging broader domain knowledge.

\subsection{Comparison to Human-Constructed KG}
To further validate the quality of our automated {\ours} paper deconstruction, we manually construct KGs for several papers and compared them against the nodes generated by {\ours}. Recognizing that manual annotation is a time-consuming process requiring significant domain expertise, we randomly selected three papers from the {\ours} corpus and invited expert PhD candidates to annotate them. 

As shown in Table \ref{tab:kg_comparison}, the high weighted F1-scores demonstrate that our automated KG construction closely aligns with human annotation. Crucially, manually annotating a single paper takes 30-60 minutes, which highlights the significant efficiency gains of our automated pipeline.

\subsection{Analysis of Paper Represent Granularity}
\label{app:granularity}
Our hierarchical graph design (Section \ref{sec:design}) is intentionally flexible, built to preserve a paper's natural methodological structure rather than forcing a uniform method decomposition. While most papers (71\%) in {\ours} are decomposed into 2-5 techniques, a significant portion of papers are represented as single, atomic nodes (12\%) or complex works with 6+ nodes (17\%), ensuring that potentially useful information from high-level methodologies to fine-grained details is captured at its appropriate level.

\section{Analysis on Target Paper}
\label{app:app_experiments}

\begin{figure}[!htbp]
  \centering
  \includegraphics[width=0.9\linewidth]{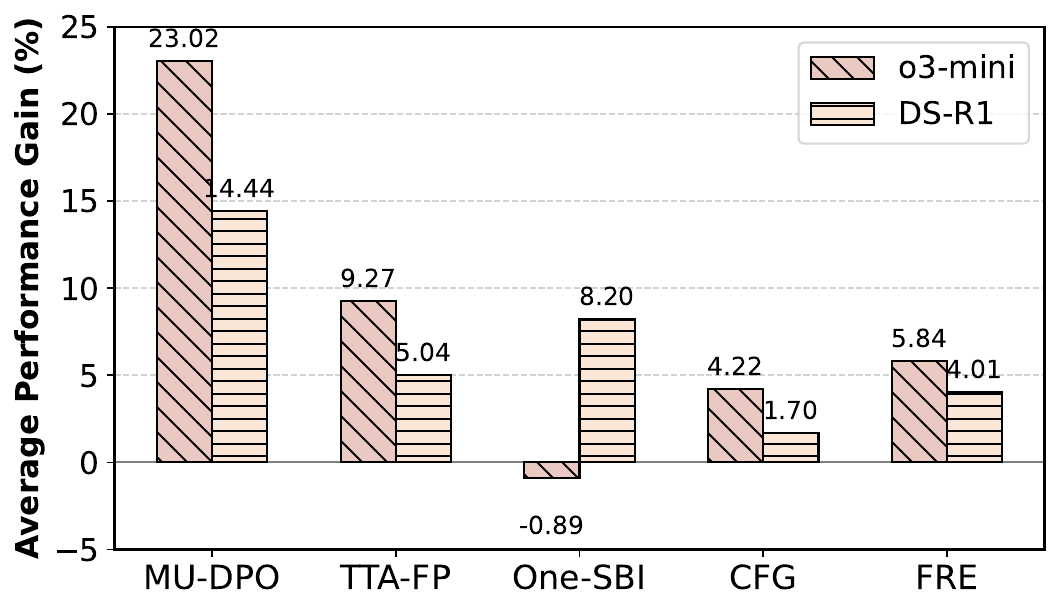}
  \caption{Average performance gain per paper.}
  \label{fig:gain}
\end{figure}

As illustrated in Figure~\ref{fig:gain}, the effectiveness of {\ours} is highly contingent on the target paper, with performance occasionally degrading. 
Bad cases stem from two primary failure modes: 
(1) \textbf{Over-reliance on retrieved code}, where the agent prioritizes generic snippets over the paper's unique implementation details; 
and (2) \textbf{Over-focus on core components}, where excelling at core techniques highlighted by {\ours} leads to the neglect of secondary objectives.

More fundamentally, this performance disparity is tied to the paper's research archetype. 
\textit{analytical papers}, such as MU-DPO\cite{DBLP:conf/icml/LeeBPWKM24}, which synthesize and refine existing techniques, benefit substantially as their components are well-represented in {\ours}. 
Conversely, \textit{methodological papers} like One-SBI\cite{DBLP:conf/icml/GlocklerDWWM24}, which introduce fundamentally novel architectures, find less directly applicable knowledge, as their core innovations have limited precedent in the corpus. 
This outcome is logical, as the performance bottleneck shifts from \textit{knowledge augmentation} to the intrinsic \textit{innovative capacity} of the base LLM itself.

\begin{table*}[!htp]
    \centering
    \small 
    \renewcommand{\arraystretch}{1.2}
    \begin{tabularx}{\textwidth}{l X c} 
    \toprule[1.2pt]
    \textbf{Domain} & \textbf{Subfield} & \textbf{Count} \\
    \midrule
    Generative AI & Diffusion Models, Distribution Modeling, Controllable Generation & 14 \\
    AI Safety \& Robustness & Test-Time Adaptation, Continual Learning, Toxicity Detection & 13 \\
    Simulation-Based Inference & Inverse Inference, Bayesian Inference, Scientific Computing & 8 \\
    Reinforcement Learning & Exploration Strategies, Agent Decision-making, Policy Optimization & 7 \\
    Mechanistic Interpretability & Circuit Analysis, Feature Visualization, Geometric Analysis & 7 \\
    \bottomrule[1.2pt]
    \end{tabularx}
    \caption{Domains of the original 42 papers in xKG. The sum of counts exceeds the total number of papers due to interdisciplinary classifications.}
    \label{tab:xkg_domain}
\end{table*}
\begin{table*}[!htp]
    \centering
    \small 
    \renewcommand{\arraystretch}{1.2}
    \begin{tabularx}{\textwidth}{l X c} 
    \toprule[1.2pt]
    \textbf{Domain} & \textbf{Subfield} & \textbf{Count} \\
    \midrule
    Generative AI & Diffusion Models, Distribution Modeling, Controllable Generation & 18 \\
    AI Safety \& Robustness & Test-Time Adaptation, Continual Learning, Toxicity Detection & 15 \\
    Simulation-Based Inference & Inverse Inference, Bayesian Inference, Scientific Computing & 8 \\
    Mechanistic Interpretability & Circuit Analysis, Feature Visualization, Geometric Analysis & 8 \\
    Reinforcement Learning & Exploration Strategies, Agent Decision-making, Policy Optimization & 7 \\
    Model Adaptation \& Efficiency & Transfer Learning, Visual Prompting, Model Reprogramming & 4 \\
    \bottomrule[1.2pt]
    \end{tabularx}
    \caption{Updated domains of the 56 papers in the expanded xKG+, where some papers span multiple domains.}
    \label{tab:xkg_domain_updated}
\end{table*}

\begin{table*}[!htp]
    \centering
    \small
    \renewcommand{\arraystretch}{1.2}
    \begin{tabularx}{\textwidth}{
        >{\raggedright\arraybackslash\hsize=1.0\hsize}X
        >{\centering\arraybackslash\hsize=0.7\hsize}X
        >{\centering\arraybackslash\hsize=0.7\hsize}X
        >{\centering\arraybackslash\hsize=1.3\hsize}X
        >{\centering\arraybackslash\hsize=1.3\hsize}X
    }
    \toprule[1.2pt]
    \textbf{Setting} & \textbf{Method} & \textbf{Model} & \textbf{bridging-data-gaps} & \textbf{sample-specific-masks} \\
    \midrule
    \textbf{w/o xKG} & BasicAgent & \texttt{\texttt{o3-mini}} & 11.55 & 24.09 \\
    \textbf{xKG (42 papers)} & BasicAgent & \texttt{o3-mini} & 13.93 & 31.73 \\
    \textbf{xKG+ (56 papers)} & BasicAgent & \texttt{o3-mini} & 44.64 & 42.47 \\
    \bottomrule[1.2pt]
    \end{tabularx}
    \caption{Replication scores (\%) under different settings on the \textit{bridging-data-gaps} and \textit{sample-specific-masks} tasks.}
    \label{tab:xKG-results}
\end{table*}

\begin{table*}[!htp]
    \centering
    \small 
    \renewcommand{\arraystretch}{1.2}
    
    \begin{tabular*}{\textwidth}{@{\extracolsep{\fill}} cccccc @{}}
    \toprule[1.2pt]
    \textbf{Paper} & \textbf{HumanKG Nodes} & \textbf{xKG Nodes} & \textbf{Precision (w)} & \textbf{Recall (w)} & \textbf{F1-Score (w)} \\
    \midrule
    CFDG & 4 & 3 & 1.00 & 0.90 & 0.95 \\
    RND  & 11 & 7 & 0.92 & 0.86 & 0.89 \\
    TENT & 4 & 4 & 1.00 & 1.00 & 1.00 \\
    \bottomrule[1.2pt]
    \end{tabular*}
    
    \caption{Comparison of automatically-constructed xKG and human-constructed KG.}
    \label{tab:kg_comparison}
\end{table*}
\begin{table*}[!htp]
    \centering
    \small
    \renewcommand{\arraystretch}{1.2}
    \begin{tabularx}{\textwidth}{l l X}
    \toprule[1.2pt]
    \textbf{Paper} & \textbf{Domain} & \textbf{Subfield} \\
    \midrule
    \textbf{FRE} & Reinforcement Learning
        & Zero-Shot RL, Unsupervised Pre-training, Functional Reward Encoding, Transformer-VAE, Offline RL \\
    \textbf{TTA-FP} & Efficient Machine Learning
        & Test-Time Adaptation, Backpropagation-Free Optimization, CMA-ES, Prompt Tuning, Quantized Model Deployment \\
    \textbf{MU-DPO} & AI Safety \& Alignment
        & Mechanistic Interpretability, Direct Preference Optimization, Toxicity Reduction, Activation Intervention, SVD Analysis \\
    \textbf{One-SBI} & Scientific Machine Learning
        & Simulation-Based Inference, Amortized Bayesian Inference, Diffusion Models, Transformer with Structured Attention, Score-Matching \\
    \textbf{CFG} & Natural Language Processing
        & Controllable Text Generation \\
    \bottomrule[1.2pt]
    \end{tabularx}
    \caption{Domains and subfields of the PaperBench tasks evaluated in this work.}
    \label{tab:paperbench-domain}
\end{table*}
\begin{table*}[!htp]
    \centering
    \small
    \renewcommand{\arraystretch}{1.2}
    \begin{tabularx}{\textwidth}{l X c}
    \toprule[1.2pt]
    \textbf{Our Abbr.} & \textbf{PaperBench Name} & \textbf{CodeDev Nodes} \\
    \midrule
    \textbf{FRE} \cite{DBLP:conf/icml/FransPAL24} 
        & \textit{fre}
        & 306 \\
    \textbf{TTA-FP} \cite{DBLP:conf/icml/NiuMCWZ24}
        & \textit{test-time-model-adaptation}
        & 86 \\
    \textbf{MU-DPO} \cite{DBLP:conf/icml/LeeBPWKM24} 
        & \textit{mechanistic-understanding}
        & 36 \\
    \textbf{One-SBI} \cite{DBLP:conf/icml/GlocklerDWWM24} 
        & \textit{all-in-one}
        & 92 \\
    \textbf{CFG} \cite{DBLP:conf/icml/SanchezSFLB24} 
        & \textit{stay-on-topic-with-classifier-free-guidance}
        & 70 \\
    \bottomrule[1.2pt]
    \end{tabularx}
    \caption{Abbreviations for the PaperBench tasks evaluated in this work.}
    \label{tab:paperbench-lite}
\end{table*}

\begin{table*}[!htp]
    \centering
    \small
    \renewcommand{\arraystretch}{1.2}
    \begin{tabularx}{\textwidth}{l c c c c c}
    \toprule[1.2pt]
    \textbf{Stage} & \textbf{LLM} & \textbf{API Calls (avg)} & \textbf{Input Tokens (avg)} & \textbf{Output Tokens (avg)} & \textbf{Cost (avg, \$)} \\
    \midrule
    Technique Extraction & o4-mini & 8.21  & 53,142.36  & 11,523.43  & 0.1092 \\
    Code Modularization  & o4-mini & 33.36 & 202,625.21 & 91,426.14  & 0.6252 \\
    Knowledge Filtering  & ---     & ---   & ---        & ---        & --- \\
    \midrule
    \textbf{Total} & o4-mini & 41.57 & 255,767.57 & 102,949.57 & 0.7344 \\
    \bottomrule[1.2pt]
    \end{tabularx}
    \caption{Average cost analysis for constructing the xKG for a single paper.}
    \label{tab:construction-cost}
\end{table*}

\begin{table*}[htp]
\centering
\small
\begin{tabular}{p{5cm}p{3.5cm}p{6cm}}
\toprule
\textbf{Hyperparameter} & \textbf{Value} & \textbf{Description} \\
\midrule
\multicolumn{3}{l}{\textit{\textbf{Code-RAG Module}}} \\
\texttt{code.embedder.model} & \texttt{text-embedding-3-small} & The embedding model used for code chunk vectorization. \\
\texttt{code.text\_splitter.chunk\_size} & 350 & The size of each text chunk when splitting code files. \\
\texttt{code.text\_splitter.chunk\_overlap} & 100 & The number of overlapping characters between adjacent chunks. \\
\texttt{code.retriever.faiss.top\_k} & 10 & Number of initial candidate chunks retrieved via FAISS vector search. \\
\texttt{code.retriever.llm.top\_files} & 5 & Number of top files selected by the LLM re-ranker for detailed analysis. \\
\texttt{code.exec\_check\_code} & False & A boolean flag to enable or disable the execution-based verification of generated code. \\
\bottomrule
\end{tabular}
\caption{Hyperparameters for the Code-RAG module in \ours{}.}
\label{tab:hyperparams_coderag}
\end{table*}

\begin{table*}[htp]
\centering
\small
\begin{tabular}{p{5cm}p{3.5cm}p{6cm}}
\toprule
\textbf{Hyperparameter} & \textbf{Value} & \textbf{Description} \\
\midrule
\multicolumn{3}{l}{\textit{\textbf{Paper-RAG Module}}} \\
\texttt{paper.rag} & True & A boolean flag to enable or disable the entire Paper-RAG process. \\
\texttt{paper.embedder.model} & \texttt{text-embedding-3-small} & The embedding model used for paper text vectorization. \\
\texttt{paper.text\_splitter.chunk\_size} & 350 & The size of each text chunk when splitting the paper content. \\
\texttt{paper.retriever.faiss.top\_k} & 5 & Number of relevant text excerpts retrieved from the paper via FAISS. \\
\bottomrule
\end{tabular}
\caption{Hyperparameters for the Paper-RAG module in \ours{}.}
\label{tab:hyperparams_paperrag}
\end{table*}

\begin{table*}[htp]
\centering
\small
\begin{tabular}{p{5cm}p{3.5cm}p{6cm}}
\toprule
\textbf{Hyperparameter} & \textbf{Value} & \textbf{Description} \\
\midrule
\multicolumn{3}{l}{\textit{\textbf{Knowledge Graph Retrieval}}} \\
\texttt{retrieve.embedding\_model} & \texttt{all-MiniLM-L6-v2} & The sentence-transformer model used for calculating similarity between techniques. \\
\texttt{retrieve.technique\_similarity} & 0.6 & The minimum similarity score required for a technique to be retrieved from the KG. \\
\texttt{retrieve.paper\_similarity} & 0.6 & The minimum similarity score required for a paper to be retrieved from the KG. \\
\bottomrule
\end{tabular}
\caption{Hyperparameters for Knowledge Graph retrieval.}
\label{tab:hyperparams_kg_retrieval}
\end{table*}

\begin{table*}[htp]
\centering
\small
\begin{tabular}{p{5cm}p{3.5cm}p{6cm}}
\toprule
\textbf{Hyperparameter} & \textbf{Value} & \textbf{Description} \\
\midrule
\multicolumn{3}{l}{\textit{\textbf{Global \& Model Profile Configuration}}} \\
\texttt{log\_level} & DEBUG & Sets the verbosity of logging. \\
\texttt{kg\_path} & \texttt{storage/kg} & The directory where the constructed Knowledge Graph is stored. \\
\texttt{max\_prompt\_code\_bytes} & 52100 & The maximum size in bytes for code content included in a prompt to the LLM. \\
\texttt{model} & \texttt{DeepSeek-V3} & The primary foundation model for the agent's base tasks. \\
\texttt{paper\_model} & \texttt{o4-mini} & A specialized model used specifically for extracting and rewriting techniques from papers. \\
\texttt{code\_model} & \texttt{o4-mini} & A specialized model used for rewriting and debugging code. \\
\bottomrule
\end{tabular}
\caption{Common global settings and an example model profile (\texttt{basic-deepseek-v3}). Specific models can be defined for different sub-tasks, allowing for flexible and optimized model selection.}
\label{tab:hyperparams_global}
\end{table*}

\clearpage
\onecolumn
\section{Running Examples of xKG}
\label{app:xkg-case}
\begin{figure}[!htbp]
  \centering
  \includegraphics[width=0.9\linewidth]{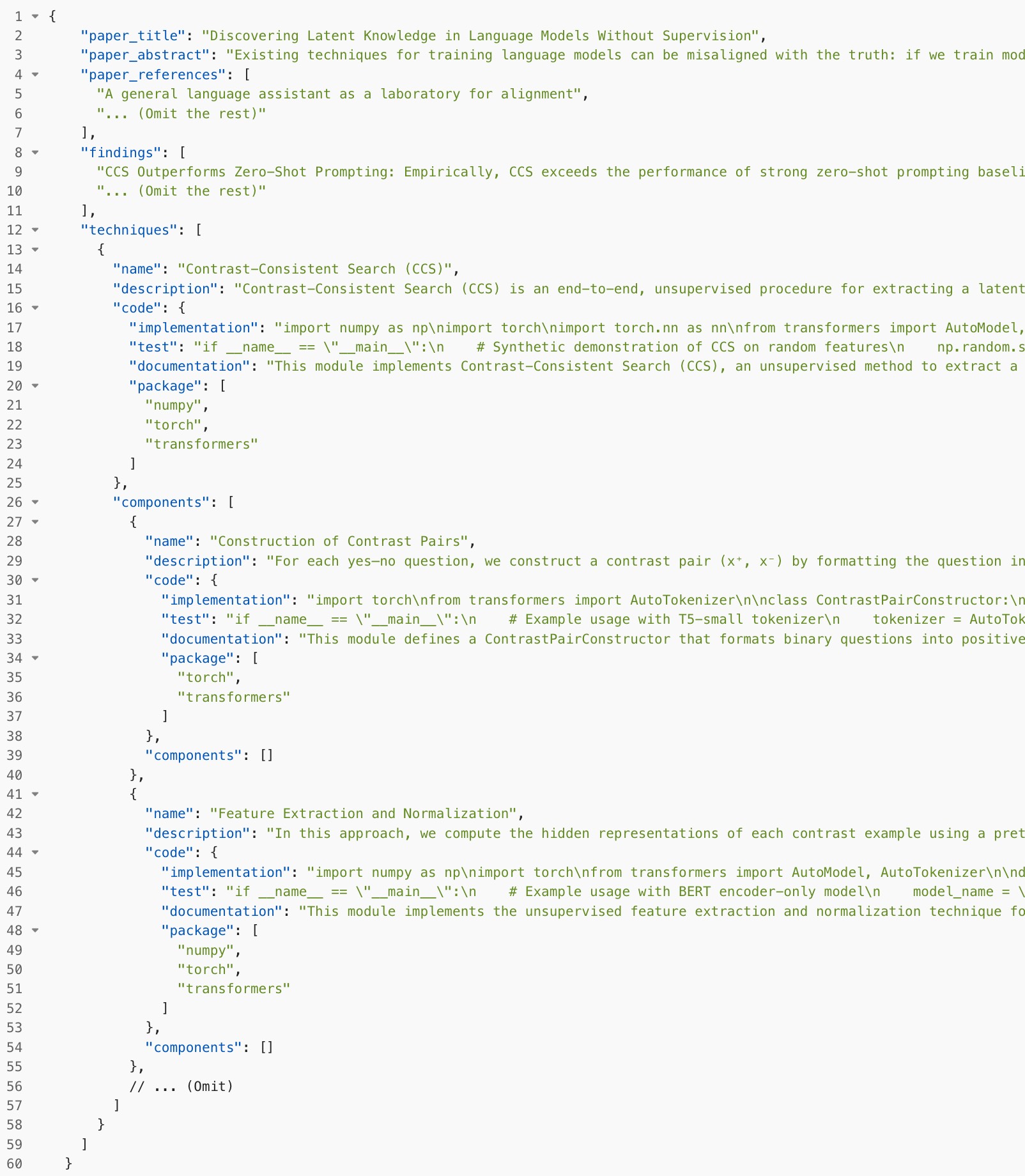}
  \caption{An example of structural {\ours} data storage. Paper Nodes are stored as JSON files, with technique and code nodes embedded as structured dictionaries, where key-value pairs are used to create a one-to-one mapping representing the implementation relationship.}
  \label{fig:example1}
\end{figure}

\clearpage

\onecolumn
\newpage
\section{Prompts}
\label{app:prompts_details}

In this section, we showcase some of the key prompts used in the full pipeline of our system, which serve as a reference. 
The prompts are organized by their functional role in the pipeline: paper parsing, code repository parsing, and knowledge graph construction.

\subsection{Paper Parsing}

\begin{tcolorbox}[
    enhanced,
    title=Prompt for Extracting References from .bbl File,
    breakable
]
\# Task

You are provided with a .bbl file \{bbl\}. 
Please extract the titles of all the references in the .bbl file.
\newline

\# Output

1. Output the extracted reference titles in the form of a string list.

2. If no reference is available, please return None.
\newline

Please wrap your final answer between two \texttt{\textasciigrave\textasciigrave\textasciigrave} in the end.
\end{tcolorbox}

\begin{tcolorbox}[
    enhanced,
    title=Prompt for Extracting Paper Contributions,
    breakable
]
\# Task

You are provided with the paper titled \{title\}. Here are the main sections of the paper: \{sections\}. Furthermore, key equations from the paper are provided to help you understand its specific algorithms: \{equations\}.
Your task is to analyze the provided research paper and identify its \textbf{Core Components}. For each Component, you must provide a clear, concise, and implementable definition.
\newline

\# INSTRUCTIONS

1.  \textbf{Identify Core Components}: Read the paper to identify its primary components. A componnet is not limited to a single algorithm; 
it can be a novel methodology, reusable techniques, key insight/finding, open-source datasets/benchmarks, etc.

2.  \textbf{Categorize Each Component}: Assign one of the following types to each component you identify:
    \begin{itemize}
        \item \texttt{Methodology}: A novel, end-to-end procedure proposed by the paper for solving a problem. This can be an entire algorithm or model architecture design that addresses a specific research challenge. It must correspond to a systematic and complete end-to-end code implementation. When composed of multiple atomic sub-techniques, represent using the "components" field. Ensure the methodology can be implemented standalone, instead of a generic theoretical definition or a high-level outline of a framework.
        \item \texttt{Technique}: A self-contained and algorithmically implementable component, applied within the paper's Methodology or Experiment Process. It is either a novel module from this work, or a traceable technique from prior research. When composed of multiple atomic sub-techniques, represent using the "components" field. Ensure each technique can be implemented standalone, requiring NO integration with other modules to constitute a single code module. Exclude theoretical points and experimental tricks not directly tied to code implementation. Move them to the "Finding" category.
        \item \texttt{Finding}: A significant empirical or theoretical insight which can refer to an intriguing experimental finding, a powerful theoretical proof, or a promising research direction.
        \item \texttt{Resource}: A PUBLICLY available dataset or benchmark originally constructed in this paper.
    \end{itemize}

3.  \textbf{Define and Detail}: For each component, provide a detailed definition adhering to the following rules:
    \begin{itemize}
        \item \textbf{Fidelity}: All definitions must originate strictly from the provided paper. Do not invent details.
        \item \textbf{Atomicity \& Modularity}: Each component, whether high-level or a component, should be defined as a distinct, self-contained unit. Explain its inputs, core logic, and outputs.
        \item \textbf{Reproducibility}: Retain as much original detail as possible. The definition should be comprehensive enough for an engineer or researcher to understand and implement it.
        \item \textbf{Structure}: If a `Methodology` or a `Technique` is composed of smaller `Technique`s, represent this hierarchical relationship using nested bullet points. This is crucial for understanding how the parts form the whole. Don't list techniques individually if they're already part of a larger technique/methodology.
    \end{itemize}

\# OUTPUT FORMAT

Organize the extracted techniques into a list of dictionaries, with the final answer wrapped between two \texttt{\textasciigrave\textasciigrave\textasciigrave} markers. The keys for each dictionary are described below:

1. name: str, the name of the component, expressed as a concise and standardized academic term, intended to precisely capture its core identity while facilitating efficient indexing and retrieval from other literature.

2. type: str, One of `Methodology`, `Technique`, `Finding`, or `Resource`.

3. description: str, A detailed, self-contained explanation of the component, focusing on what it is, how it works, and its purpose. For implementable items, describe the whole process without missing any critical steps and implementation details. For insights, describe the core discovery. Maximize the retention of description and implementation details from the original text.

4. components: List[dict], Optional, If the component is a complex `Methodology` or `Techinque` composed of multiple smaller techniques, this field lists its key sub-techniques. Each sub-technique listed here must also be defined separately as a complete technique object following this same JSON schema (with `name`, `type` and `description` as dictionary keys), allowing for hierarchical and recursive decomposition. ATTENTION: Only `Methodology` and `Technique` can have `Technique` as its components!!! 
\newline

Now please think and reason carefully, and wrap your final answer between two \texttt{\textasciigrave\textasciigrave\textasciigrave} in the end.
\end{tcolorbox}

\subsection{Code Repository Parsing}

\begin{tcolorbox}[
    enhanced,
    title=Prompt for Generating Code Repository Overview,
    breakable
]
\# Task

Analyze this GitHub repository \{name\} and create a structured overview of it.
\newline

\# Input

1. The complete file tree of the project:
\{file\_tree\}

2. The README file of the project:
\{readme\}
\newline

\# Output

Create a detailed overview of the project, including:

1.Overview (general information about the project)

2.System Architecture (how the system is designed)

3.Core Features (key functionality)

Organize the overview in a clear and structured markdown format.
\newline

Please wrap your final answer between two \texttt{\textasciigrave\textasciigrave\textasciigrave} in the end.
\end{tcolorbox}

\begin{tcolorbox}[
    enhanced,
    title=Prompt for Finding Associated Paper from Code,
    breakable
]
\# Task

Analyze this GitHub repository \{name\}, and determine whether this repository is directly associated with a specific academic paper.
\newline

\# Input

The README file of the project:
\{readme\}
\newline

\# Output

1. If you can find clear evidence that this repository is the official or direct code implementation of a specific academic paper, return the full title of the paper as a string.

2. If there is no sufficient evidence to identify a directly corresponding paper (e.g., only general descriptions, multiple papers, or no paper mentioned), return None.
\newline

Please wrap your final answer between two \texttt{\textasciigrave\textasciigrave\textasciigrave} in the end.
\end{tcolorbox}

\subsection{Knowledge Graph Construction}

\begin{tcolorbox}[
    enhanced,
    title=Prompt for Rewriting a Technique's Description,
    breakable
]
\# Task

Your task is to refine and enhance the description of a technical concept extracted from a research paper \{paper\}. The goal is to produce a clear, concise, and comprehensive description that accurately captures the essence of the technique.
\newline

\# Input

1.  Technical Concept from the paper \{paper\}: 
\{technique\}

2. Relevant Excerpt of this Technique:
\{excerpt\}
\newline

\# Output

Return a precise and comprehensive description, presented as a single, continuous paragraph written in a comprehensive, academic style. Avoid using bullet points, numbered lists, or other form of itemization.

1. Ensure the technique precisely matches the original description. DO NOT alter, expand, or reduce the scope of the technique. Ignore other related techniques and only FOCUS ON this technique.

2. Strictly adhering to the original description, augment its implementation details based on the provided excerpts. All formulas, parameter configurations, and implementation details must be extracted from the given excerpts, ensuring strict adherence to them. Avoid any summarization, inference, or omission.

3. If the excerpts offer no new information, leave the description unchanged. Your response MUST be based solely on the original description and provided excerpts. The inclusion of ANY external information or fabricated details is strictly forbidden!!!

4. Ensure that the provided description is precise, complete, and possesses sufficient detail to correspond to a specific implementation.
\newline

Now please think and reason carefully, and wrap your final answer between two \texttt{\textasciigrave\textasciigrave\textasciigrave} in the end.
\end{tcolorbox}

\begin{tcolorbox}[
    enhanced,
    title=Prompt for Identifying Relevant Code Snippets,
    breakable
]
\# Task

Your task is to analyze a list of code files retrieved from a GitHub repository, and identify which files are directly relevant to the implementation of a specific technical concept defined in an academic paper \{paper\}.
\newline

\# Input

1.  Technical Concept Definition from the paper \{paper\}: 
\{technique\}

2. Overview of the Code repository:
\{overview\}

3. Relevant Code Files:
\{file\_snippets\}
\newline

\# Output

Return a list of filenames formatted as ["xx", "xx", ...], sorted in \textbf{descending} order of relevance of the technical concept.

1. Exclude any file not DIRECTLY correspond to the concrete implementation and configurarion of this technique (e.g., tests, documentation, other technique implementation).

2. Confirm that a direct implementation exists within your provided file list. If no such implementation can be found, return None.

3. Return the filename list even if there's only one file. 
\newline

Now please think and reason carefully, and wrap your final answer between two \texttt{\textasciigrave\textasciigrave\textasciigrave} in the end.
\end{tcolorbox}

\begin{tcolorbox}[
    enhanced,
    title=Prompt for Reranking Retrieved Techniques,
    breakable
]
\# Task

Your task is to analyze a list of technique implementations retrieved from the knowledge base, and identify which techniques are directly relevant to the implementation of a specific technical concept.
\newline

\# Input

1. Technical Concept Definition: 
\{technique\}

2. Relevant Technique implementations:
\{relevant\_techniques\}
\newline

\# Output

Return a list of (technique\_name, apply\_guidance) tuples formatted as [("", ""), ("",""), ...], sorted in descending order of relevance to the technical concept. 
The guidance should be a short explanation of how the technique applies to the current scenario and what modifications are needed for adaptation. Use clear and definite wording, avoiding parentheses.

1. Exclude any techniques not relevant to the concrete implementation of this technique. 

2. Ensure the returned technique name exactly matches the original one.

3. For technologies with identical core definitions, keep the one whose application is most relevant.

4. If no such technique can be found, return None.

5. Return the filename list even if there's only one relevant technique. 
\newline

Now please think and reason carefully, and wrap your final answer between two \texttt{\textasciigrave\textasciigrave\textasciigrave} in the end.
\end{tcolorbox}

\begin{tcolorbox}[
    enhanced,
    title=Prompt for Rewriting Code for a Leaf Technique,
    breakable
]
\# Task

Your task is to transform a collection of disparate source code snippets, which are the official implementation of a technique component from a research paper \{paper\}, into a single, self-contained, and executable code block. The final code block must be clean, well-documented, and easy for others to understand and run.
\newline

\# Input

1. Abstract of the paper \{paper\}:
\{abstract\}

2. Technical Concept Definition from the paper \{paper\}:
\{technique\}

3. Relevant Code Files:
\{file\_snippets\}
\newline

\# Workflow

1. Analyze: Understand the technique's inputs, outputs and workflow from the paper. Focus ONLY on THIS technique, ignoring the mentioned context and related techniques.

2. Isolate \& Extract: Based on the description of the technique, determine what is its PRECISE role and functionality, and extract ONLY the code you identified as belonging to \{technique\}. Other mentioned associated techniques \textbf{MUST BE IGNORED AND EXCLUDED}.

3. Refactor: Integrate the extracted code by removing hard-coded values, isolating the core algorithm, and standardizing it with proper documentation and type hints.

4. Assemble \& Test: Build the final script and add an test block as a runnable example. Ensure accuracy and conciseness, avoiding unnecessary output.

5. Documentation: Write a brief and concise documentation of the code logic, configurable options, and usage in 5-10 sentences.
\newline

\# Requirements

1. Dependency Management: Ensure all necessary imports and dependencies are included at the beginning of the code block.

2. Fidelity to the Original Technique: Strictly follow the description of the given technique to organize the code. ONLY focus on the implementation that DIRECTLY corresponds to THIS technique!!! (e.g., if the technique is a loss function definition, implement only the code for its calculation. Ignore all other parts of the algorithm's implementation, even if provided in the code snippets.)

3. Code Encapsulation and Documentation:
\begin{itemize}
    \item Encapsulate the core logic of the technique into one or more functions/classes.
    \item Every function and class method must include a comprehensive docstring explaining its purpose, parameters, and return values.
    \item All function arguments and return values must have clear type hints.
    \item Preserve original parameters and comments from the source code.
\end{itemize}
4. Reproducibility and Testing:
\begin{itemize}
    \item A main execution block, starting with the comment \texttt{\# TEST BLOCK}, is required at the end of the file, which serves as a practical usage example and a test case.
    \item The test case should use parameters from the code repository or paper. If missing, create and state your own defaults.
\end{itemize}
5. Fidelity to the Original Logic: 
\begin{itemize}
    \item You must strictly adhere to the algorithmic logic present in the provided code snippets. Your role is to refactor and structure, not to re-implement or invent new logic.
    \item Minimal, necessary modifications are permitted (e.g., renaming variables for clarity, adapting function signatures for dependency injection), but the core computational steps must remain identical to the original author's implementation.
\end{itemize}
6. Documentation of Usage Scenarios: Provide a concise and fluent document of the code module’s core logic, configurable options, and usage. Limit the description to 5-10 clear and coherent sentences.
\newline

\# Output

1. Implement the technique standalone without relying on external, undefined components. Return an executable code block and a corresponding documentation, each wrapped between two \texttt{\textasciigrave\textasciigrave\textasciigrave} .

Example:

[... Reasoning Steps ...]

\texttt{\textasciigrave\textasciigrave\textasciigrave}python

[... Core Implementation of the technique ...]

[... Ignore other relevant techniques ...]

\# TEST BLOCK
[... Example Usage ...]

\texttt{\textasciigrave\textasciigrave\textasciigrave}

The brief documentation of the code:

\texttt{\textasciigrave\textasciigrave\textasciigrave}

[...Brief Documentation ...]

\texttt{\textasciigrave\textasciigrave\textasciigrave}

2. Verify that the generated code does not exceed the scope of the technique's definition. If the technique requires integration with other modules to constitute a single code module, return None. If no direct implementation of the technique is found in the given code snippets, also return None.
\newline

Now, please proceed with the task, following the workflow and adhering to all requirements. Generate the final code block and documentation wrapped between two \texttt{\textasciigrave\textasciigrave\textasciigrave} separately at the end.
\end{tcolorbox}

\begin{tcolorbox}[
enhanced,
title=Prompt for Rewriting Code for a Composite Technique,
breakable
]
\# Task

Your task is to transform a collection of disparate source code snippets, which are the official implementation of a technique component from a research paper {paper}, into a single, self-contained, and executable code block. The final code block must be clean, well-documented, and easy for others to understand and run.
\newline

\# Input

Abstract of the paper \{paper\}:

\{abstract\}

Technical Concept Definition from the paper \{paper\}:

\{technique\}

Sub-techniques and Associated Code:

\{sub\_techniques\}

Relevant Code Files:

\{file\_snippets\}
\newline

\# Workflow

Analyze: Understand the technique's inputs, outputs and workflow from the paper.

Locate: Fully reuse the code of the provided sub-techniques. For any uncovered parts, locate the relevant implementation logic from the given code snippets.

Refactor: Integrate the extracted code by removing hard-coded values, isolating the core algorithm, and standardizing it with proper documentation and type hints.

Assemble \& Test: Build the final script and add an test block as a runnable example. Ensure accuracy and conciseness, avoiding unnecessary output.

Documentation: Write a brief and concise documentation of the code logic, configurable options, and usage in 5-10 sentences.
\newline

\# Requirements

Dependency Management: Ensure all necessary imports and dependencies are included at the beginning of the code block.

Fidelity to the Original Technique: Strictly follow the description of the given technique to organize the code. ONLY focus on the implementation that DIRECTLY corresponds to THIS technique!!! (e.g., if the technique is a loss function definition, implement only the code for its calculation. Ignore all other parts of the algorithm like model definition or training loop). Return None if no direct implementation is found.

Code Encapsulation and Documentation:
\begin{itemize}
\item Encapsulate the core logic of the technique into one or more functions/classes.
\item Every function and class method must include a comprehensive docstring explaining its purpose, parameters, and return values.
\item All function arguments and return values must have clear type hints.
\item Preserve original parameters and comments from the source code.
\end{itemize}
Reproducibility and Testing:
\begin{itemize}
\item A main execution block, start with the comment \texttt{\# TEST BLOCK}, is required at the end of the file, which serves as a practical usage example and a test case.
\item The test case should use parameters from the code repository or paper. If missing, create and state your own defaults.
\end{itemize}
Fidelity to the Original Logic:
\begin{itemize}
\item You must strictly adhere to the algorithmic logic present in the provided code snippets. Your role is to refactor and structure, not to re-implement or invent new logic.
\item Minimal, necessary modifications are permitted (e.g., renaming variables for clarity, adapting function signatures for dependency injection), but the core computational steps must remain identical to the original author's implementation.
\end{itemize}
Documentation of Usage Scenarios: Provide a concise and fluent document of the code module’s core logic, configurable options, and usage. Limit the description to 5-10 clear and coherent sentences.
\newline

\# Output

1. Implement the technique standalone without relying on external, undefined components. Return an executable code block and a corresponding documentation, each wrapped between two \texttt{\textasciigrave\textasciigrave\textasciigrave}.

Example:

[... Reasoning Steps ...]

\texttt{\textasciigrave\textasciigrave\textasciigrave}python

[... Core Implementation of the technique ...]

[... Ignore other relevant techniques ...]

\# TEST BLOCK

[... Example Usage ...]

\texttt{\textasciigrave\textasciigrave\textasciigrave}

The brief documentation of the code:

\texttt{\textasciigrave\textasciigrave\textasciigrave}

[...Brief Documentation ...]

\texttt{\textasciigrave\textasciigrave\textasciigrave}

2. Verify that the generated code does not exceed the scope of the technique's definition. If the technique requires integration with other modules to constitute a single code module, return None. If no direct implementation of the technique is found in the given code snippets, also return None.
\newline

Now, please proceed with the task, following the workflow and adhering to all requirements. Generate the final code block and documentation wrapped between two \texttt{\textasciigrave\textasciigrave\textasciigrave} separately at the end.
\end{tcolorbox}

\begin{tcolorbox}[
enhanced,
title=Prompt for Verifying Rewritten Code,
breakable
]
\# Task

Your task is to determine if the given code block strictly follows the provided technique description and relevant code files.
\newline

\# Input

Technical Concept Definition from the paper \{paper\}:
\{technique\}
Relevant Code Files:
\{file\_snippets\}
Implemented Code Block:
\{code\}
\newline

\# Output

1.Return False if the implementation is unrelated to the technique.

2.Return False if the implementation contains core logic cannot be located in the given relevant code files.

3.Return False if the implementation contains logics not covered in the technique description (e.g., the technique defines a submodule, but the code implements the full algorithm).

4.Return True if the code implements exactly what is specified in the technique description without adding any unnecessary features beyond the technical concept, and strictly follows the implementation in the given code files.
\newline

Now please think and reason carefully, provide a detailed analysis process for the above criteria, and wrap your final answer between two \texttt{\textasciigrave\textasciigrave\textasciigrave} in the end.
\end{tcolorbox}
\begin{tcolorbox}[
enhanced,
title=Prompt for Decomposing a Task into Techniques,
breakable
]
\# Task

Your task is to decompose a complex academic task into its automic fundamental techniques based on its description.
\newline

\# Input

Academic Task Definition:
\{description\}
\newline

\# Output

Return a list of (name, description) tuples in the format [("...", "..."), ("...", "...")], sorted by their importance to the task composition in descending order. Use clear and definite wording, avoiding parentheses.
Each tuple must represent a distinct, fundamental academic concept that is reusable and traceable in other literature.
Each tuple is explicitly mentioned or directly relevant to the target task.
Avoid overly broad or vague techniques; each should have a clear, specific code implementation. Avoid trivial techniques like Cosine Similarity that require no literature review.
If the task's implementation does not involve any specific academic concepts (e.g., purely engineering, configuration, or organizational task), simply return None.
\newline

Now please think and reason carefully, and wrap your final answer between two \texttt{\textasciigrave\textasciigrave\textasciigrave} in the end.
\end{tcolorbox}

\end{document}